%% file: main.tex
\documentclass[twocolumn,10pt]{asme2ej} 

\usepackage[utf8]{inputenc}
\usepackage[T1]{fontenc}

\usepackage{graphics}
\usepackage{epsfig}
\usepackage{xcolor}
\usepackage{colortbl}
\usepackage{cite}
\usepackage{hyperref} 
\usepackage{soul}	

\usepackage{amsmath} 
\DeclareMathOperator{\sgn}{sgn}
\usepackage[super, comma, sort&compress]{natbib} 

\providecommand{\keywords}[1]
{
  \textbf{\textit{Keywords---}} #1
}

\title{\LARGE \bf Configuration and Fabrication of Preformed Vine Robots}

\author{Nathaniel Agharese$^{1}$ and Allison M. Okamura$^{1}$
\thanks{$^{1}$N. Agharese and A. M. Okamura are with the Department of Mechanical Engineering, Stanford University, Stanford, CA 94305 USA.\newline
(e-mail: \{agharese, aokamura\}@stanford.edu).}%
}

\begin{document}

\maketitle
\thispagestyle{empty}
\pagestyle{empty}

\begin{abstract}

Vine robots are a class of soft continuum robots that grow via tip eversion, allowing them to move their tip without relying on reaction forces from the environment. Constructed from compliant materials such as fabric and thin, flexible plastic, these robots are able to grow many times their original length with the use of fluidic pressure. They can be mechanically programmed/preformed to follow a desired path during growth by changing the structure of their body prior to deployment. We present a model for fabricating preformed vine robots with discrete bends. We apply this model across combinations of three fabrication methods and two materials. One fabrication method, taping folds into the robot body, is from the literature. The other two methods, welding folds and connecting fasteners embedded in the robot body, are novel. Measurements show the ability of the resulting vine robots to follow a desired path and show that fabrication method has a significant impact. Results include bend angles with as little as 0.12 degrees of error, and segment lengths with as low as 0.36 mm of error. The required growth pressure and average growth speed of these preformed vine robots ranged from 11.5 to 23.7kPA and 3.75 to 10 cm/s, respectively. These results validate the use of preformed vine robots for deployment along known paths, and serve as a guide for choosing a fabrication method and material combination based on the specific needs of the task.

\end{abstract}


\input{Introduction}

\input{Fabrication}


\input{Modeling}

\input{Methods}

\input{Results}

\section{Conclusion}
Preformed vine robots are a promising alternative to traditionally complex robots for precise navigation of known environments up to varying lengths. The goal of this work was to characterize the accuracy to which we can fabricate preformed vine robots with discrete bends, and to assess the growth performance of the resulting vine robots. We presented a model for fabricating these robots from a set of DH parameters defining a desired shape at complete eversion. We showed that for the methods and materials tested the method of fabrication has a significant impact on the accuracy of each DH parameter, the material has no significant impact, and repetitive growth has a significant impact on link twist and joint angles. One of our novel methods, the loop method, performs significantly better than the others in matching  link twist, but significantly worse than the others in matching link length.

Future work involves improving the model for fabrication using the loop method to make the link length accuracy on par with the accuracy of the other DH parameters. There are also limits to joint angles in terms of accuracy and the ability to grow that were not within the scope of this work. Investigating the lower limits of link lengths until joint interference becomes too severe would enhance our ability to apply our model to a broader range of growth paths and path planners.



\section*{Acknowledgements}

The authors thank Iain Darby of Dounreay for introducing the need for iterative preformed vine robots, and Tenzing Joshi and Ren Cooper of Lawrence Berkeley National Lab for collaborations and guidance on vine robots for nuclear inspection.

\section*{Author Disclosure Statement}

No competing financial interests exist.

\section*{Funding Information}

This work was performed under the auspices of the U.S. Department of Energy by Lawrence Berkeley National Laboratory under Contract DE-AC02-05CH11231. The project was funded by the U.S. Department of Energy, National Nuclear Security Administration, Office of Defense Nuclear Nonproliferation Research and Development (DNN R\&D), and the United States Federal Bureau of Investigation contract 15F06721C0002306.

\bibliographystyle{unsrtnat}
\bibliography{references}

\end{document}

%% file: Introduction.tex
\section{Introduction}
The continuum, length changing, and compliant characteristics of soft growing robots are well suited for deployment in large, complex, unknown, or sensitive operations such as inspection~\cite{Takeichi2017IROS, dong2017development}, manipulation~\cite{McMaha2006ICRA, hannan2003kinematics}, and minimally invasive surgery~\cite{abidi2018highly, Burgner2015ToR}. This work focuses on one class of soft growing continuum robots: vine robots~\cite{HawkesScienceRobotics2017}.
Vine robots have potential applications such as sensor deployment~\cite{GruebeleRoboSoft2021}, archaeological research~\cite{CoadRAM2020, GanRAL2020}, teleoperated manipulation~\cite{StroppaICRA2020}, endovascular surgery~\cite{Mingyuan2021MedicalRobotics}, and communications~\cite{BlumenscheinRAL2018}. Such applications require knowledge of tip pose, typically achieved with a combination of sensors and a kinematic model. Vine robot kinematics are determined by their internal pressure, interactions with the environment, and steering inputs. These steering inputs are either active and reversible, or mechanically pre-programmed via preforming.


\begin{figure}
    \centering
    \vspace{2mm}
    \includegraphics[width=0.9\columnwidth]{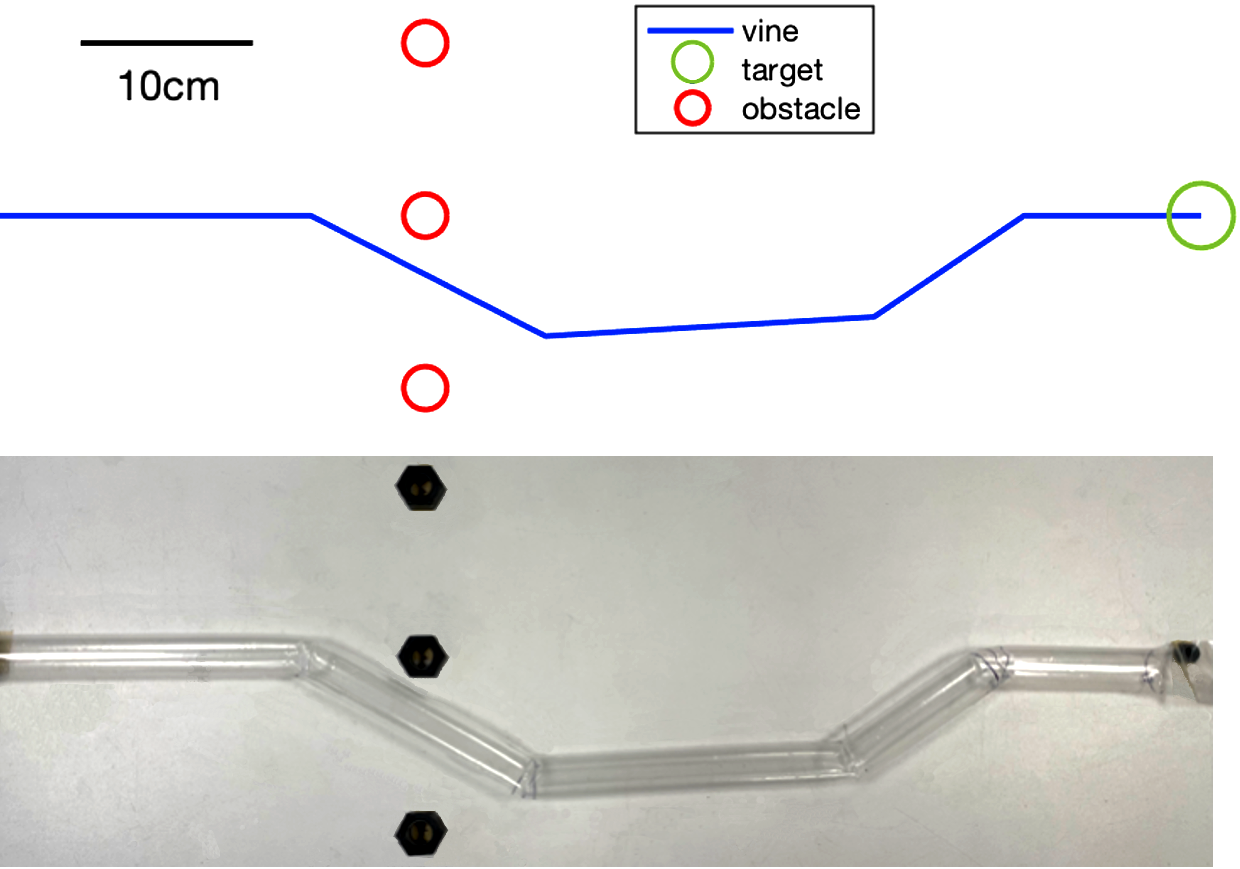}
    \caption{We tested preformed vine robots in 3D space using two materials and three fabrication methods. \textbf{Top:} An example of a desired path for a preformed vine robot to grow around obstacles and reach a target tip pose. \textbf{Bottom:} An implementation of this path with a preformed vine robot that uses low-density polyethylene as the material and tape adhesive as the fabrication method, designed via our new fabrication model.}
    \vspace{-5mm}
    \label{fig:glamour}
\end{figure}

Active steering methods are useful for exploring unknown spaces and typically operate by shortening the entire length of the robot using pneumatic or tendon-based actuators that are controlled with a human user~\cite{GreerSoRo2019, BlumenscheinROBOSOFT2018}. Deriving the tip pose of soft continuum robots with these steering methods typically involves kinematic models with a piece-wise constant curvature (PCC) assumption~\cite{Jones2006ToR, GreerICRA2017}. However, these models and their accompanying actuation methods are limited when it comes to applications in long and highly tortuous paths. Furthermore, active steering methods lack the precision to avoid obstacles along the entire length of the robot, such as what is achieved in Figure \ref{fig:glamour}.

Jitosho et al.~\cite{jitosho2023passive} and Wang et al.~\cite{wang2020dexterous} address some of these shortcomings by designing vine robots with shape locking capabilities. Jitosho et al. passively lock the shape of the vine robot during growth, while Wang et al. actively lock the robot shape independent of growth. Both systems suffer from an exponential increase in control effort with actuated length. Hawkes et al.~\cite{HawkesScienceRobotics2017} present a method of non-reversible actuation during deployment by releasing pre-constrained material along one side of the vine robot. This method is complex and non-reversible without deconstruction and reconstruction of the robot, resulting in a large amount of pre-deployment effort for a single growth. Preforming vine robots can reduce preparation time and duration of deployment, while still enabling the avoidance of obstacles along the entire length.

When the environment is known, preformed vine robots are able to overcome the modeling and precision limitations of vine robots that use active steering methods. Discrete shapes have been achieved by enforcing pinches of material along the length of the vine robot~\cite{GreerIJRR2020}, and continuum shapes have been achieved by heat shrinking low-density polyethylene (LDPE) around a mold~\cite{AghareseICRA2018, SladeIROS2017} or combining a series a pinches to create a continuous curvature~\cite{BlumenscheinROBOSOFT2018}. Reducing environmental interactions using preformed vine robots requires accurate kinematic modeling, but does not require sensors because these robots are unable to change their shape in real time and they tend to be deployed without strict growth control~\cite{AghareseICRA2018, BlumenscheinRAL2018, BlumenscheinTRO2022, BlumenscheinROBOSOFT2018, GreerIJRR2020, SladeIROS2017}. Blumenschein et al. demonstrate a kinematic model for a preformed vine robot with discrete folds~\cite{BlumenscheinROBOSOFT2018}. However, their model is for continuum shapes that are generated using a series of very small, discrete bends. Additionally, existing work with preformed vine robots lacks a thorough investigation of path matching and growth behaviors as a function of preforming method and robot material~\cite{AghareseICRA2018, BlumenscheinRAL2018, BlumenscheinTRO2022, BlumenscheinROBOSOFT2018, GreerIJRR2020, SladeIROS2017}.


The contributions of this work are two novel methods for preforming vine robots in 3D space, a model for fabricating preformed vine robots to achieve a desired shape with large discrete bends, and an experimental assessment of growth and shape matching capabilities of the resulting vine robots.

%% file: Fabrication.tex
\section{Fabrication Methods} \label{implementation}

We use two common vine robot materials: 75 micron thick low-density polyethylene (LDPE) and 182.5 micron thick thermoplastic polyurethane (TPU) coated ripstop nylon~\cite{HawkesScienceRobotics2017, BlumenscheinROBOSOFT2018, CoadRAM2020}. We refer to the TPU-coated ripstop nylon material as `fabric' in this paper. We use these materials to create the inflatable beam that becomes the vine robot body. For the LDPE, we heat seal two ends of a tube of LDPE whose diameter equates to the inflated diamter of the vine robot. For the fabric, we use a sheet of ripstop nylon with TPU-coating on one side. We ultrasonically weld two lines on the TPU-coated side together to create the tube that becomes the vine robot body. This ultrasonic welding technique is further discussed in Section \ref{welding}.

After creating the inflatable beam, we fabricate vine robots using three methods to create discrete bends along the length of the robot body. We (1) secure folds in the body using tape adhesive, (2) secure folds using ultrasonic welding, or (3) connect two points on the body together using rigid connectors and a loop attachment technique. These methods are described in more detail in the following sections.

\begin{figure}
    \centering
    \includegraphics[width=\columnwidth]{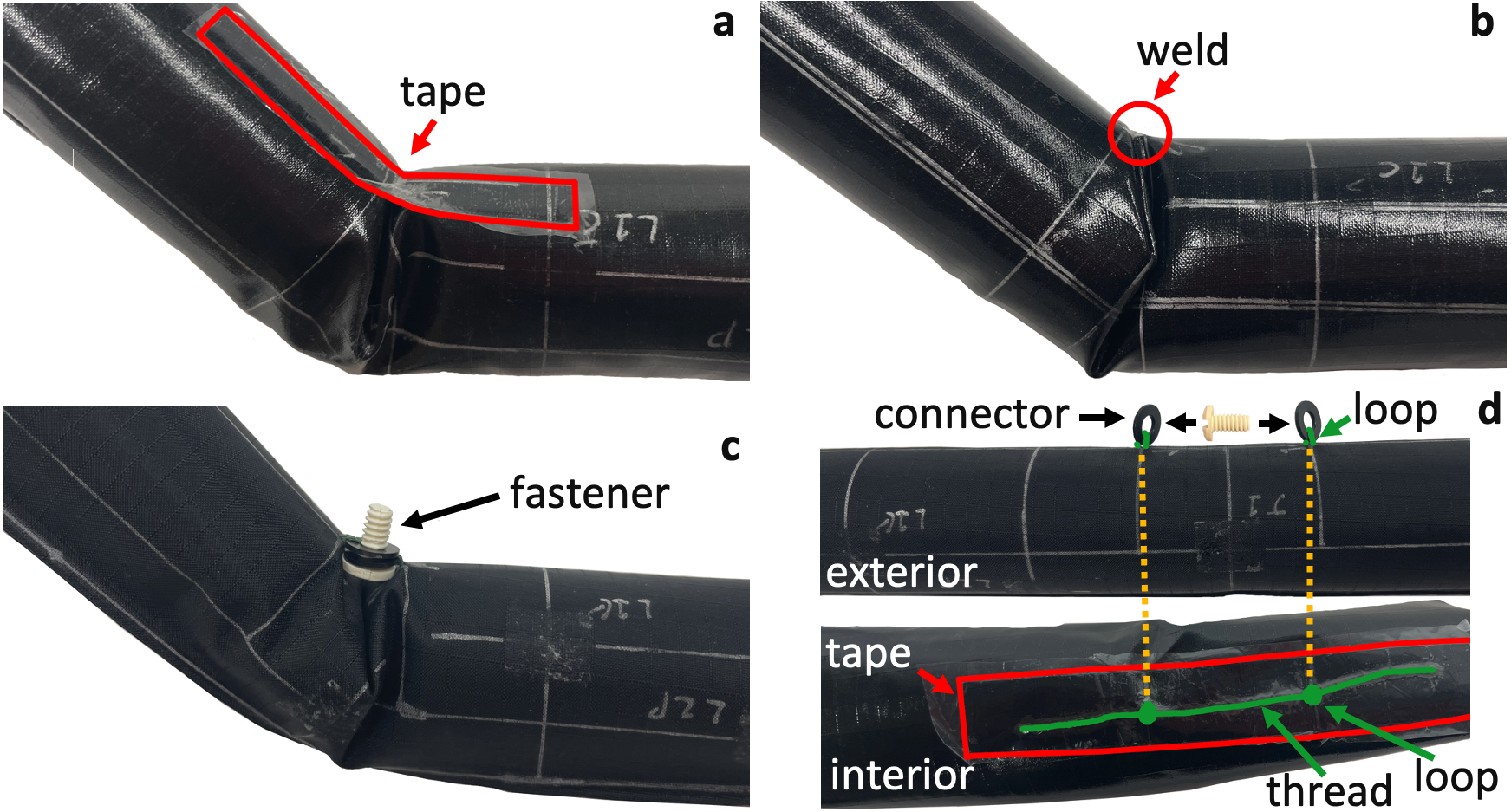}
    \caption{Examples of bends fabricated into a preformed vine robot made of TPU-coated ripstop nylon using each of the methods discussed in Section \ref{implementation}. \textbf{(a)} Adhesive tape secures folds in the material. \textbf{(b)} Ultrasonic welding secures folds in the material. \textbf{(c)} Two rigid connectors are brought together by a fastener to create a bend. \textbf{(d)} Views of the exterior and interior of the robot where the rigid connectors are attached to the material with thread and tape before being secured together by a fastener.}
    \label{fig:manufacturing}
\end{figure}

\subsection{Adhesive Tape}
\label{tape}

Adhesive tape has previously been used to secure folds in a vine robot body to create discrete and continuous bends in preformed vine robots~\cite{BlumenscheinROBOSOFT2018, GreerIJRR2020}. We apply this method to serve as a control against the novel methods described in the following two sections. We use portions of tape that are 60 mm long and 6 mm wide to create these bends. The vine robots constructed from fabric have a TPU coating only on one side. Preliminary testing showed that our adhesive tape (MD 9000 Marker Attachment Tape, Marker-Tape) adhered better to this TPU coated side than the non-coated side. So, as shown in Figure \ref{fig:manufacturing}(a), we place the tape on this side for vine robots created with fabric. Our adhesive tape is double-sided, so we cover the exposed surface with a thin layer of LDPE to prevent self adhesion.

This method is useful for quickly creating bends, but the tape is susceptible to peeling away after repeated growths or increasing vine robot body pressure. The critical failure mode of this method is the tape separating from the material. This failure is easily remedied by recreating the fold and attaching a new strip of tape. Additionally, this method is a poor approximation of bringing two points on the surface of the vine together as described in the model in Section \ref{inverse_kin}. The width of the tape translates to connecting two lines together, in contrast to two points. This violates the point assumption in our model, and would theoretically result in joint angles with larger magnitudes.

\subsection{Ultrasonic Welding}
\label{welding}

Ultrasonic welding is a type of friction welding; a process that bonds materials together with a combination of lateral force and friction generated heat~\cite{besharati2014advances}. Specifically, ultrasonic welding uses a combination of pressure and high frequency, low amplitude, vibrations~\cite{bhudolia2020advances}. This technique is well suited for vine robots made of TPU-coated ripstop nylon. It can also be applied to LDPE, but we found that it resulted in a large material fail-rate for the thickness we were using.

Similar to the tape method (Section \ref{tape}), we create folds in the vine robot body, but we instead bind these folds via ultrasonic welding (Figure\ref{fig:manufacturing} (b)). Additionally, when manufacturing with TPU-coated fabric, we create our folds with the TPU coating on the exterior of the body because the TPU-on-TPU welds are stronger than welding the non-TPU coated side to itself. This method creates a more permanent bend compared to the tape method. However, this requires more careful fabrication because we cannot undo and remake the fold. And while folds made from ultrasonic welding tend to fail at higher pressures than folds made with tape, the failure mode is a tear in the material that is irreparable. This is because the connection that creates a bend only involves the vine robot material, not an adhesive like tape. The only way for the bend to come undone is for the bond of the material to come undone, or tear. 

\subsection{Loops}
\label{Loops}
The `loop' method attaches small, rigid connectors into the material of the vine robot body by looping high-tension thread through these connectors and the body (Figure \ref{fig:manufacturing}(d)). These connectors are then fastened together to create a bend, as shown in Figure \ref{fig:manufacturing}(c) We use nylon washers with an outer diameter of 9 mm and an inner diameter of 4.4 mm as the connectors. We attach them to the material of the robot body with a 0.9 mm needle and micro-filament thread rated at 222 N (Microfilament Braided Line, Power Pro Spectra). This attachment involves looping the thread through the connector and the robot body material two times. This redundancy reduces slack when the connector is pulled away from the body by adding capstan friction to prevent slipping.


After attaching the connectors, we secure the thread to the interior surface of the robot body using adhesive tape as shown in Figure \ref{fig:manufacturing}(d). The distinction of `interior' vs `exterior' surfaces is only relevant for vine robots made from TPU-coated fabric. We selected the TPU-coated surface to be the interior of the vine robot because that surface is better for tabe adhesion as described in Section \ref{tape}. The tape is 13 mm wide and approximately 5 mm longer than the length of thread it covers. The use of adhesive tape seals the holes in the robot body created during the attachment of the connectors, and restricts movement of the thread and the connectors when in tension. This adhesion is further strengthened when the vine robot body is pressurized because the pressure pushes the tape against the body.

These steps can be done for an arbitrary number of connectors at once as long as they are in series along the length of the vine. In this work, we apply these steps to one pair of connectors at a time and leave approximately 30 mm of thread beyond the location of each connector. Figure \ref{fig:manufacturing}(d) shows a pair of connectors attached to the vine robot body, but not yet fastened together to create a bend.

We create a discrete bend by securing a pair of connectors together. In this work we use nylon screws 9.3 mm in length with a 4.5 mm major diameter and a 9.3 mm head diameter to secure a pair of connectors. The major diameter of the screw is slightly larger than the inner diameter of the connector, creating a friction joint similar to that of a press fit. We found this to be sufficient to hold a pair of connectors together at operating pressures. This process can also be undone to disconnect the pair of connectors and undo the bend as shown in Figure \ref{fig:manufacturing}(d). While it can be undone, the position of the bend cannot be adjusted like with the tape adhesive method (Section \ref{tape}).

Preformed vine robots are typically restricted in their range of deployable environments because they always grow into one specific shape. This fabrication method overcomes that limitation because we can create and remove bends between deployments to change the shape that the vine robot grows into. This method is also more time intensive than the tape method, and has a similar fabrication time to the ultrasonic welding method. It also adds more material to the vine robot body than the tape or welding methods, increasing the resistance to growth.

%% file: Modeling.tex
\section{Modeling} \label{Model}


\begin{figure}[t!]
    \centering
    \includegraphics[width = \columnwidth]{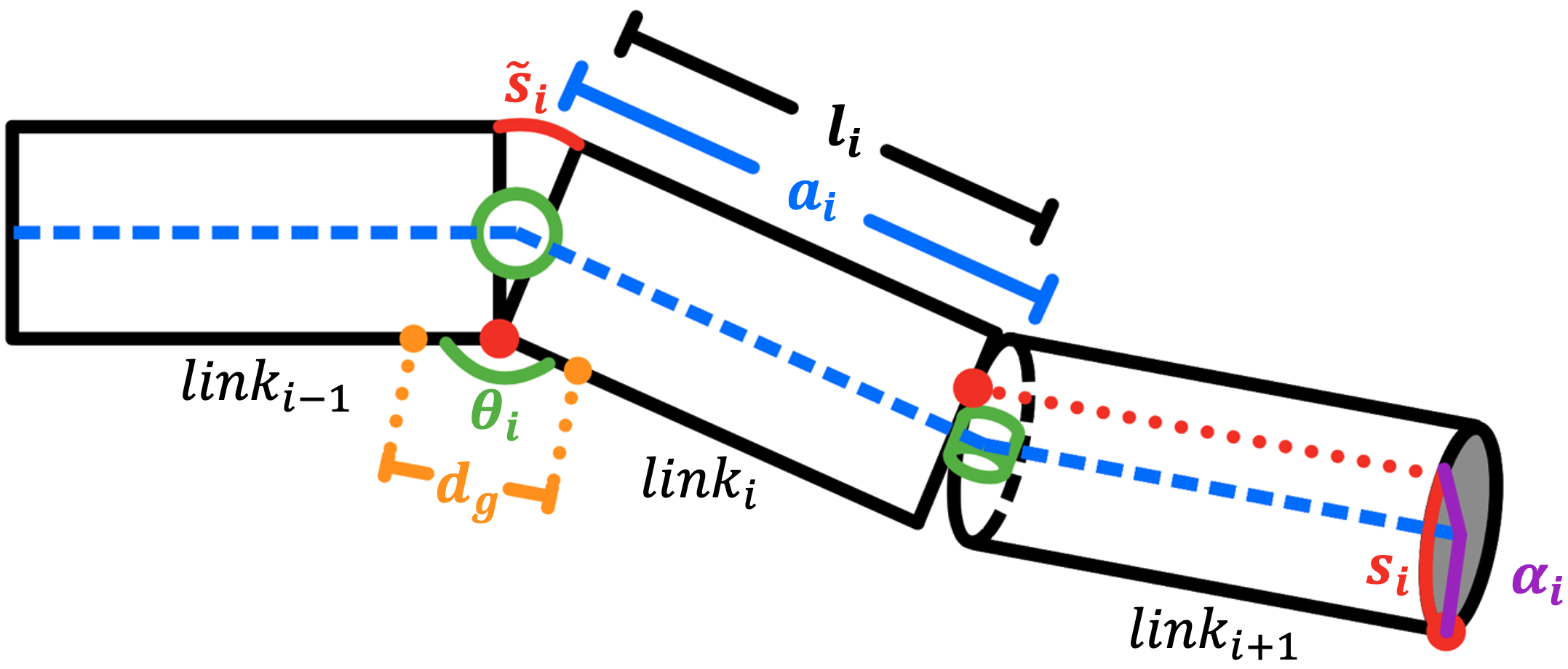}
    \vspace{0.2pt}
    \caption{An overlay of the DH parameter and cylindrical representations of the vine robot. The DH parameter elements, defined in Section \ref{Kinematic Model}, are in blue (links), green (joints), and purple (link twist). The cylindrical elements, defined in Section \ref{inverse_kin}, are in black (cylinders) and red (joints and their fabrication parameters). The points that are brought within a known distance of each other instead of being coincident (Section \ref{inverse_kin}) are in orange.}
    \label{fig:dh_param}
\end{figure}

Here we describe a kinematic model for fabricating a preformed vine robot with discrete bends. A common choice for modeling soft, continuum kinematics is to use a PCC representation to define Denavit-Hartenberg (DH) parameters~\cite{hannan2003kinematics, Jones2006ToR, GreerICRA2017, Della2018RoboSoft}. Since we are interested in vine robots with discrete bends, we instead use a cylindrical representation of the vine robot to define DH parameters. These parameters describe the desired shape of the fully everted robot.

\subsection{Desired Shape Model}\label{Kinematic Model}

We model the vine robot as an $R^{n}$ manipulator with $n$ joints, where the joint angles are fixed prior to eversion. The manipulator initially has one link with a joint at the base. The number of links increases by one each time a joint is everted as the robot grows. Growth distance theoretically has no impact on the joint angle for preformed vine robots. Thus, we define the desired shape of the fully everted preformed vine robot using a DH-parameter based representation.

In the DH representation, the robot comprises infinitely thin links connected by rotational joints (Fig. \ref{fig:dh_param}), with the first link connected to the world frame by a fixed joint. Each link $i$, has a link length $a_i$, link twist $\alpha_i$, joint angle $\theta_i$, and joint offset $d_i$. The link twist is demonstrated via the out-of-plane rotation of joint $i+1$ relative to joint $i$ in Figure \ref{fig:dh_param}. The joint offset $d_i$ is always zero for this model. This representation reflects the configuration of the vine robot at complete eversion, and is used to design the desired shape for an instance of growth into a given environment. However, this representation does not account for the radius of the robot or the physical location of the joints along its circumference. Both of these parameters are necessary for robot fabrication. Therefore, we define a fabrication model using a cylindrical representation of the robot and the DH parameters defined above.

\subsection{Fabrication Model}
\label{inverse_kin}
We derive fabrication parameters from desired DH parameters using a cylindrical representation of the preformed vine robot. This representation comprises a series of cylinders connected by rotational joints (Fig. \ref{fig:dh_param}). The joints are placed on the circumference of the vine robot: joint $i$ is on the circumference of cylinder $i$ and cylinder $i-1$. To create a joint, we bring two points along the length of the robot together. We define the axial distance between these points, $\Tilde{s}_i$, using the vine radius $r$ and the desired joint angle $\theta_i$ (\ref{s_til}). Each cylinder $i$ has the same radius $r$ of the inflated vine, and a unique length $l_i$. We relate $l_i$ to the link length $a_i$ using the vine robot radius $r$ and distances $\Tilde{s}_i$ and $\Tilde{s}_{i+1}$ (\ref{l_i}). An arc length $s_i$ defines the distance between joints $i$ and $i+1$ along the circumference of cylinder $i$. This arc length has a positive sense along the center axis of the vine pointing from joint $i$ to joint $i+1$. We relate $s_i$ to link twist $\alpha_i$ using the vine radius $r$ with joint angles $\theta_i$ and $\theta_{i+1}$ (\ref{s_i}). This model assumes the two points comprising a joint are separated by zero degrees along the circumference of the vine robot.
\begin{equation} \label{s_til}
    \Tilde{s}_i = 2r\theta_i
\end{equation}

\begin{equation} \label{l_i}
    l_i = a_i - \frac{\Tilde{s}_i + \Tilde{s}_{i+1}}{4}
\end{equation}

\begin{equation} \label{s_i}
    s_i = r \sgn(\theta_i \theta_{i+1}) (\alpha_i - \min(0, \pi \sgn(\theta_{i+1}))
\end{equation}

\label{two_point_gap}
Depending on the method used to connect two points, it will not possible to make them  coincident. Here, we adapt the definition of $\Tilde{s}_i$ to account for a gap of a known distance, $d_g$, between the two points of interest (\ref{s_til_gap}).

\begin{equation} \label{s_til_gap}
    \Tilde{s}_i = \frac{2d_g}{\sqrt{2 + 2\cos(\theta_i)}} + 2r\theta_i
\end{equation}

The distance $d_g$ represents the shortest distance between a pair of points used to create a joint (Fig. \ref{fig:dh_param}). When $d_g$ is zero, i.e. the two points are coincident, our model in (\ref{s_til_gap}) becomes our original model for $\Tilde{s}_i$ in (\ref{s_til}). The models defined above are for a general preformed vine robot with discrete bends, and are readily applied to such robots fabricated using the methods described in Section \ref{implementation}.

%% file: Methods.tex
\begin{figure}
    \centering
    \includegraphics[width=\columnwidth]{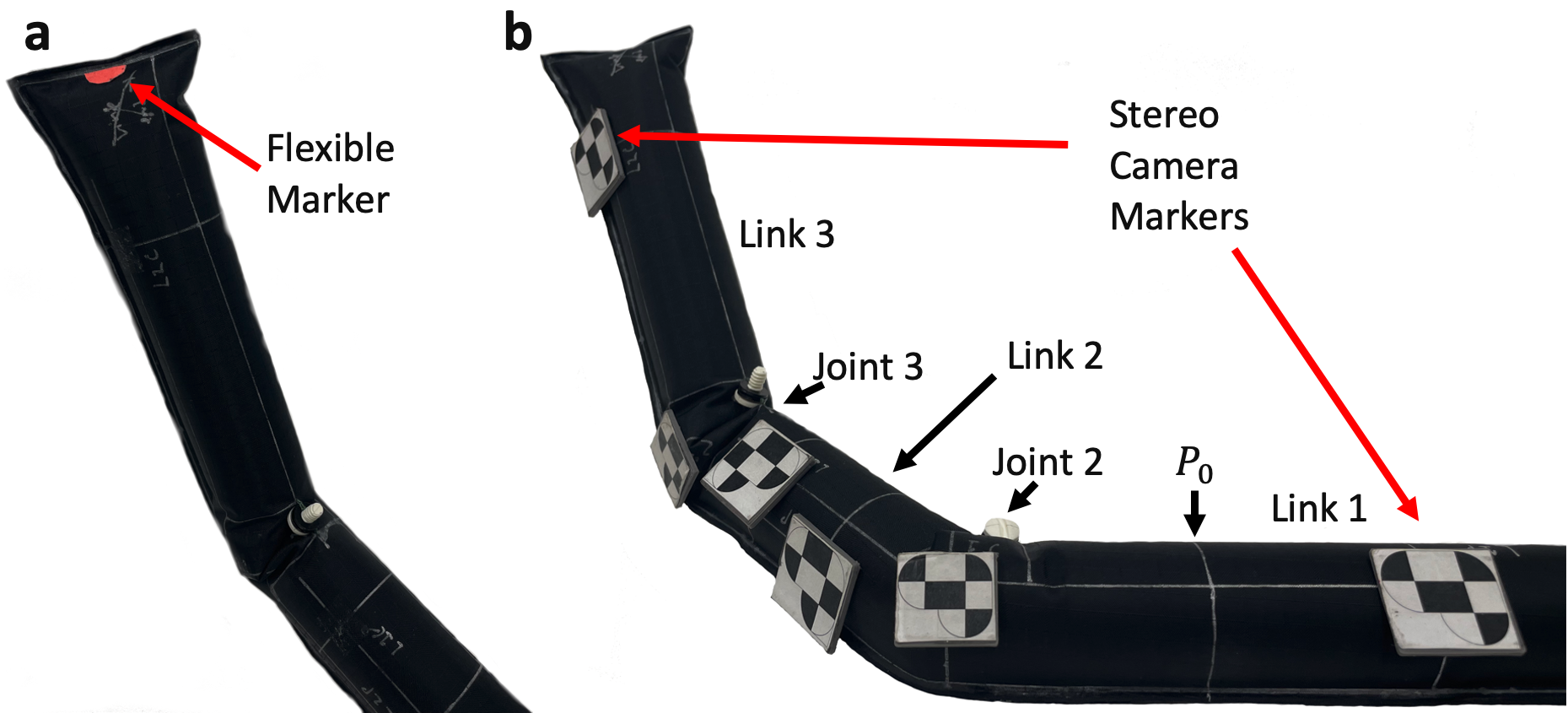}
    \caption{Rigid and flexible optical trackers are used to collected data on vine robot shape and growth behavior. (a) A flexible optical tracker is placed at the tip of the vine robot to indicate complete eversion. (b) Rigid, 6 degree-of-freedom optical markers are placed on the joints and the links of the preformed vine robot to capture its shape before and after repetitive growth.}
    \label{fig:shape_tracking}
\end{figure}

\section{Experimental methods}
\label{DH_methods}
We evaluated the DH parameter accuracy and growth characteristics of preformed vine robots made with the fabrication methods described in Section \ref{implementation} and the models in Section \ref{Model}. We controlled for the effects of material (LDPE versus fabric) by applying these techniques to both materials. We chose a three-link configuration for this assessment because that is the minimum number of links to observe the three DH parameters of interest: joint angle, link twist, and link length. All link lengths were 10 cm, joints angles were 45 degrees for links two and three, link twist was 45 degrees for link two, and all other values were zero. An example of this configuration is in Figure \ref{fig:shape_tracking}(b). We evaluated three instances of vine robots 33 mm in diameter resulting from the six combinations of fabrication method and material described in Section \ref{implementation}.

\subsection{Growth}
\label{growth assessment}
The key growth characteristics of vine robots are minimum pressure to grow and growth time, with ideal performance have the lowest possible values for both metrics. We assessed these characteristics for vine robots resulting from five of the six combinations of fabrication method and material described in Section \ref{implementation}. We excluded ultrasonic welding on LDPE because the welds consistently tore after initial pressurization. All growths started with the vine robot inverted to a point 5 cm before the first bend, we call this tip pose $P_{0}$ (Figure\ref{fig:shape_tracking}(b)). This distance allows for growth to start without interference from the folds at the joint. The vine robots had enough material before the start of the first link so they could be inverted to $P_{0}$ without internally scrunching the material. During growth, we constrained the portion of the vine robot before the start of the first link to prevent rotation or translation of the robot body that could affect our growth speed measurements. We used a closed loop pressure regulator (QB3TANKKZP6PSG, Proportion-Air, McCordsville, Indiana) and an Arduino Uno to control the pressure in the vine robot. We used a flexible optical marker at the tip of the vine robot, shown in Figure \ref{fig:shape_tracking}(a), and an LED synced to robot pressurization to record growth time via video.


After fabrication, we inverted the vine robot to $P_{0}$ and perform an initial growth. This initial growth allows the vine robot to develop the wrinkles along its length that aid in eversion for subsequent growths. Consequently, the first growth tends to require a much higher pressure than subsequent growths. To perform this initial growth, we held the vine robot to a starting pressure of 1.38 kPa until everting motion stopped. We then increased the holding pressure by 1.38 kPa and waited for everting motion to stop again. We repeated this process until the vine robot was completely everted.

\begin{figure}[t!]
    \centering
    \includegraphics[width=\columnwidth]{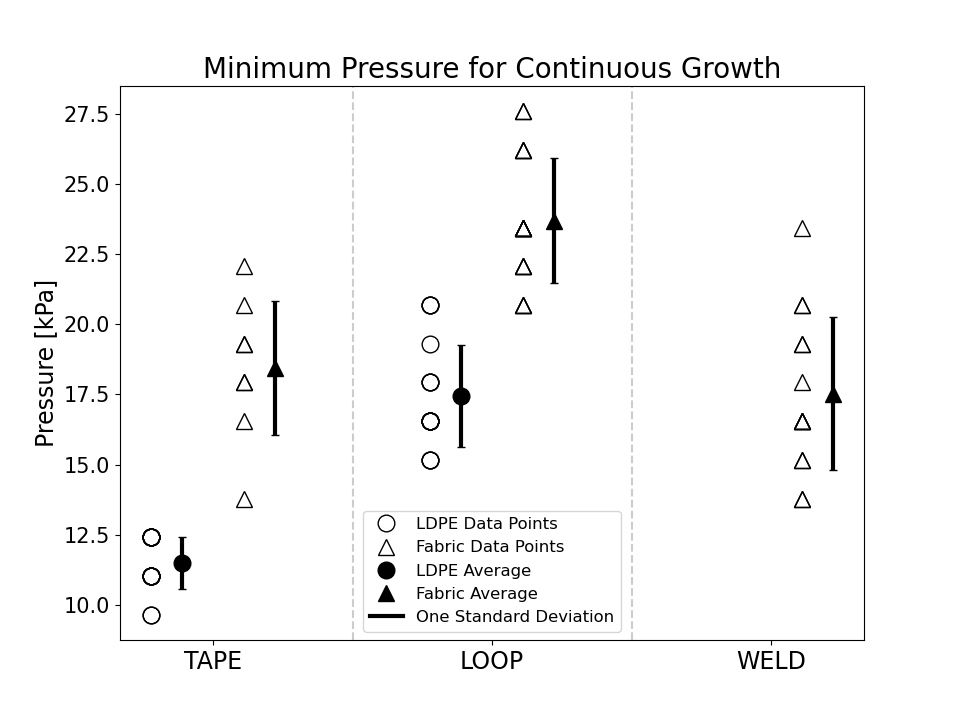}
    \caption{Minimum pressures for continuous growth. The x-axis indicates fabrication method. The marker shape indicates fabrication material: circle is LDPE and triangle is fabric.}
    \label{fig:min_pressure}
\end{figure}

After the initial growth, we used a stair casing method to iterate through pressures to identify the minimum pressure to grow for each vine robot. For each step in the staircase we inverted the vine robot to $P_{0}$, depressurized the system, changed the commanded pressure by $+/-$1.38 kPa, depending on if the previous step resulted in a growth failure/success respectively, and re-pressurized the system until the criteria for failure or success was met. A growth was a success if the vine robot grew to complete eversion without stopping. A growth was a failure if growth did not start or if everting motion stopped at any point before complete eversion. If the criteria for failure was met, we repeated the process of depressurization, increasing growth pressure, and re-pressurization until the vine robot was completely everted. These pressure increases following growth failure were not part of the stair-casing approach; they were only used to ensure each vine robot experienced 15 complete growths.

We used one of two starting pressures for this staircase test. For vine robots created with fabric using the welding method (Section \ref{welding}), fabric using the loop method (Section \ref{Loops}), or LDPE using the loop method, we used a starting pressure of 20.68 kPa. We used this starting pressure because it is in the middle of the pressure range of our pressure regulator. This range was chosen as the maximum pressure where fluctuations due to the internal control loop were an order of magnitude smaller than our desired pressure step size of 1.38 kPa. Vine robots fabricated using the tape method described in Section \ref{tape} experience complete failure of their joints at this pressure. Therefore, we used a lower starting pressure of 6.89 kPa because it is the minimum pressure for straight growth found in preliminary testing. We used two different pressures, instead of 6.89 kPa for all vine robots, so we could collect data on all fabrication method-material combinations with approximately the same number of successful growths.

We record the growth time and pressure for each successful step in the growth staircase. An LED was turned on when pressurization started to indicate the beginning of growth time. The eversion of the flexible marker shown in Figure \ref{fig:shape_tracking}(a) indicated the end of growth time.

\begin{figure} [t!]
    \centering
    \includegraphics[width=\columnwidth]{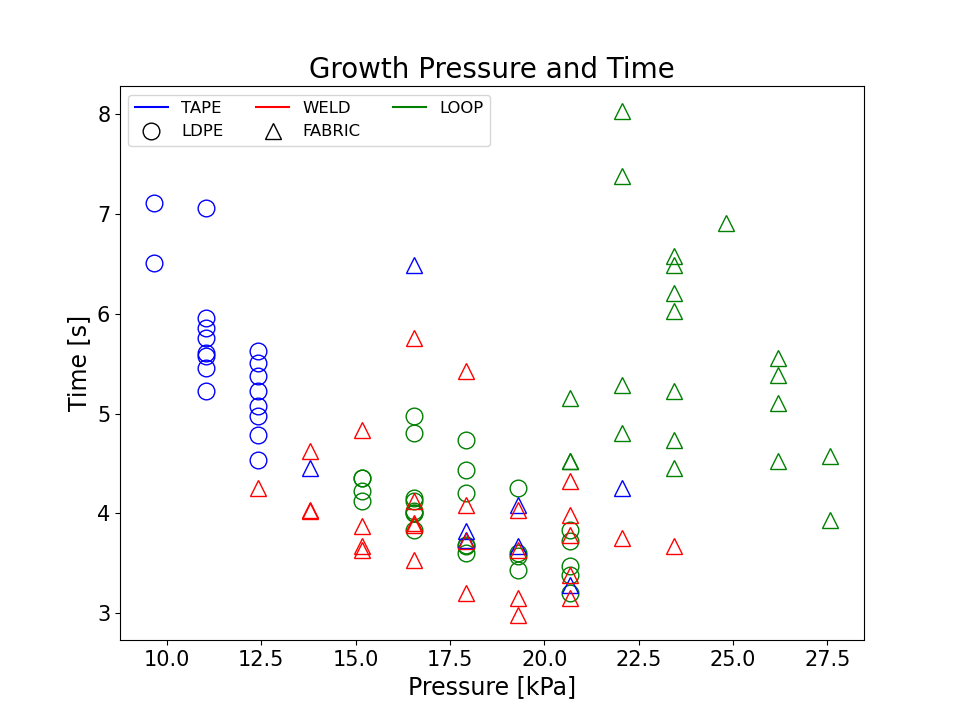}
    \caption{Growth time versus growth pressure for all preformed vine robots. The colors indicate fabrication method: tape is blue, welding is red, and loop is green. The marker shape indicates the material: circle is LDPE and triangle is fabric.}
    \label{fig:growth_times}
\end{figure}

\begin{figure*} [ht!]
    \centering
    \includegraphics[width=2\columnwidth]{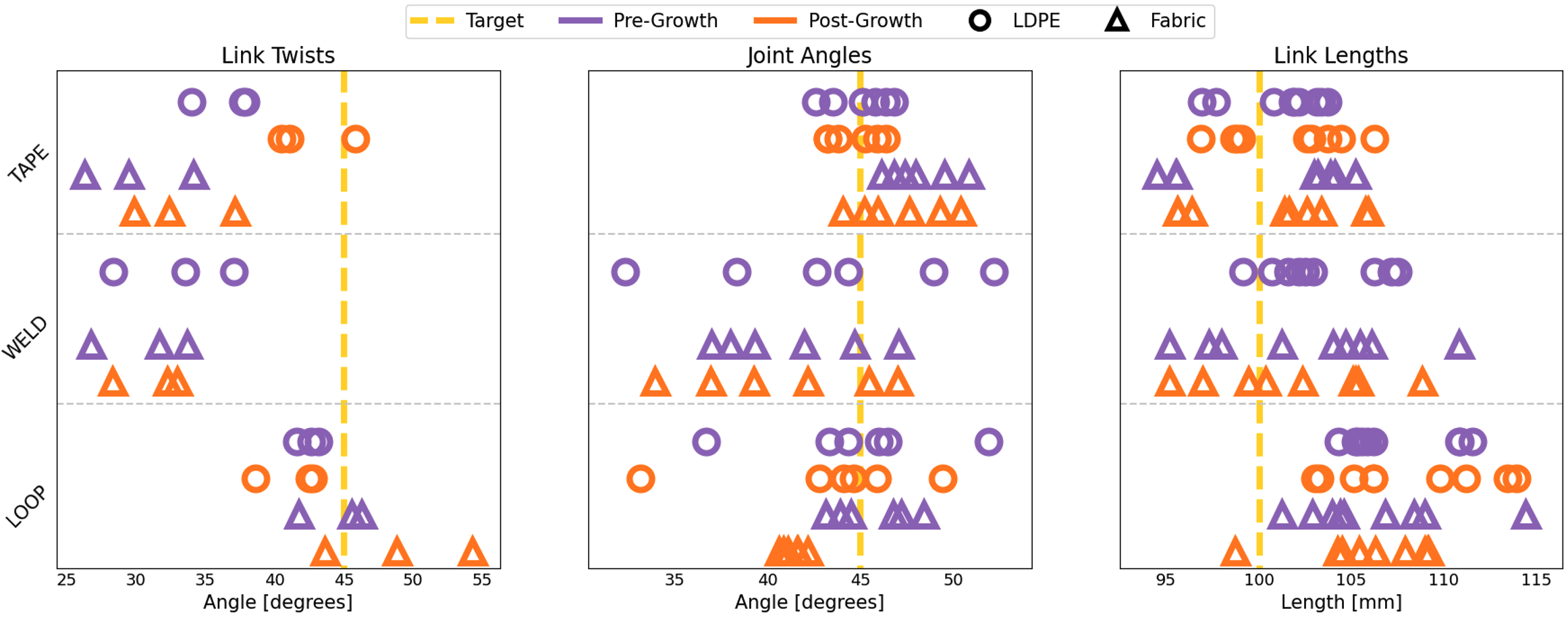}
    \caption{The DH parameters measured on each preformed vine robot. Each subplot is a DH parameter of interest. The gold line is the target value for that parameter. The y-axis indicates fabrication method. The marker shape indicates the material: circle is LDPE and triangle is fabric. The color indicates whether the data was taken before or after continuous growth: purple is before and orange is after.}
    \label{fig:dh_results}
\end{figure*}

\subsection{DH Parameter Accuracy}
We assessed DH parameter accuracy with 6-degree-of-freedom rigid optical markers and an Sx80 MicronTracker stereo camera. The markers were created by scaling MicronTracker marker templates. Each joint $i$ had three markers associated with it: a marker on the joint, a marker 7.65 cm towards joint $i-1$, and a marker 7.65 cm towards joint $i+1$. The first joint had its $i-1$ marker at the base location, and the last joint had its $i+1$ marker at the tip location. All markers for a given joint were normal to that joint axis. This distribution, shown in Figure \ref{fig:shape_tracking}(b), allowed us to measure the actual DH parameters of each vine robot.

We measured the accuracy of these parameters before and after repetitive growth. For each vine robot, we attached the rigid optical markers to their corresponding locations on the robot body. We then inflated the vine robot to its starting pressure for the staircase method in Section \ref{growth assessment} and recorded the pose data of the optical markers with the MicronTracker. The vine robot cannot invert or evert with these rigid markers attached, so we removed them after this first measurement to run the growth staircase test. After the growth test, we re-attached the markers to their same locations on the robot body. We then inflated the vine robot to the same pressure used for measurements before growth, and collected pose data via the optical trackers. This pressure was either 20.68 kPa or 6.89 kPa depending on fabrication method and material as described in Section \ref{growth assessment}. During each step of marker pose data collection, we collected 100 samples at a rate of 20 samples per second. We then averaged the raw data of these 100 samples to account for a small level of noise.



%% file: Results.tex
\label{growth_results}


\section{Results}
We measured the minimum growth pressures, growth time and pressure for each successful growth, and DH parameters for each preformed vine robot. We present a statistical analysis on the individual impact of fabrication method, material, and growth on the accuracy of vine robot DH parameters.

\begin{figure*} [ht!]
    \centering
    \includegraphics[width=2\columnwidth]{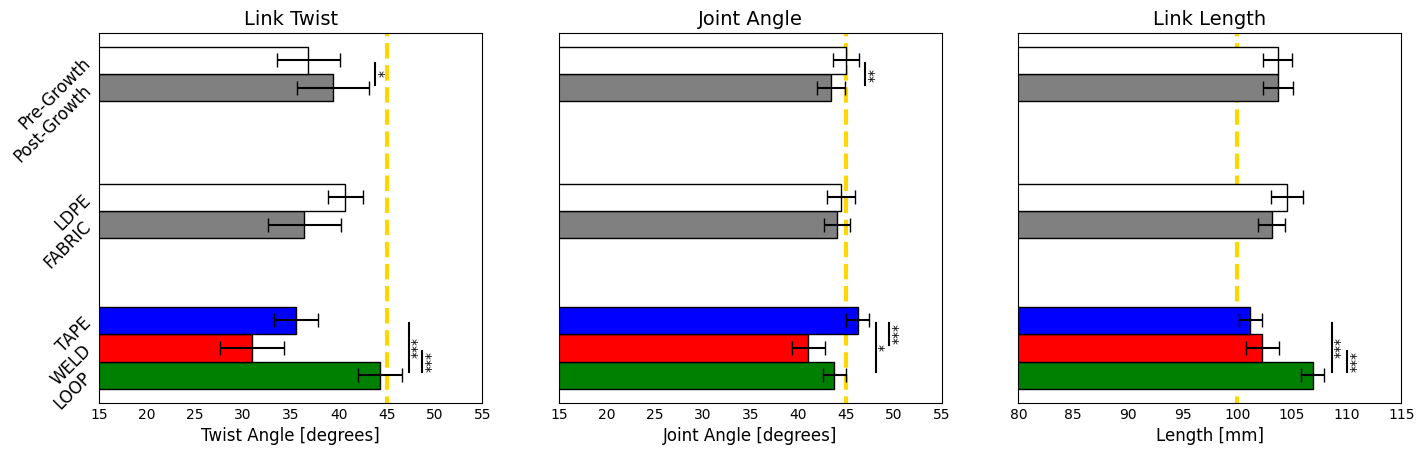}
    \caption{Each subplot shows the statistics for the data of one DH parameter. The same data is shown three times in each subplot, each time grouped by a different factor: fabrication method, material, whether measurement was before or after repetitive growth. The y-axis shows the groups within each of these factors. The fabrication method factor contains the TAPE, WELD, and LOOP groups (blue, red, and green). The fabrication material factor contains the LDPE and FABRIC groups (white and grey). The growth factor contains the Pre-Growth and Post-Growth groups (white and grey). The bars represent the mean for each group, the horizontal lines are the 95\% confidence intervals, and the vertical lines indicate group pairs that are significantly different with asterisks indicating the level of significance (*: 0.05 $>$ p $>$ 0.01, **: 0.01 $>$ p $>$ 0.001, ***: 0.001 $>$ p).}
    \label{fig:dh_stats}
\end{figure*}

\subsection{Growth}

The resulting minimum pressures for each combination of fabrication method and material are shown in Figure \ref{fig:min_pressure}. We report all local minima from the stair-casing method in Section \ref{growth assessment} for each vine. The minimum pressure for the LDPE vine robots tends to be lower than those made from fabric. This is expected because the fabric is a thicker material, and additional thickness increases the resistance to eversion at the tip. The minimum pressure for the tape method tends to be less than that of the loop method for the same material. This is also expected because the loop method adds more material to the vine robot body than the tape method, effectively further increasing its thickness. The LDPE vine robots made with the welding method have no results because they all tore open upon the first growth. The tape-fabric, loop-LDPE, and weld-fabric method-material combinations all have similar values.


The total growth time and growth pressure for all successful growths from the staircase test are shown in Figure \ref{fig:growth_times}. The method is indicated by color, and the material is indicated by marker shape. Vine robots fabricated with the tape method using LDPE have the lowest growth pressures, and one of the largest changes in growth time with pressure. Vine robots fabricated with the loop method on TPU-coated fabric have the highest growth pressures and a similar range of growth times, but the trend of growth time with pressure is not as distinct.

\subsection{DH Parameter Accuracy}

The measured DH parameters for each preformed vine robot are shown in Figure \ref{fig:dh_results}. The DH parameter is indicated at the top of each subplot, the target DH parameter value is indicated by a gold line, the fabrication method is indicated on the y-axis, the fabrication material is indicated by marker shape, and whether the measurement was taken before or after repetitive growths is indicated by color. As described in Section \ref{DH_methods}, each vine has one link twist, two joints, and three link lengths of interest (vines fabricated with welding on LDPE could not grow and therefore have no post growth DH parameters). This results in 33 link twists, 66 joint angles, and 99 link lengths.

In general, link twists for the loop method are more centered around the target value than the other methods, which are mostly below the target value. The welding method is the hardest to accurately implement, so its errors could be largely due to manufacturing error. If this is the case, more data points should reflect a larger spread in values as well. The tape method has the worst approximation of bringing two points together, which can be the cause of its DH parameter inaccuracies. Link twists also tend to be either the same or slightly larger post growth. The wrinkles around the joints might not be fully formed until repeated growths, which would result in different measured link twists and joint angles after one growth compared to after repeated growths. Joint angles for the loop and tape methods have smaller distributions than the welding method, with a few exceptions for the loop method. This is also possibly due to larger manufacturing errors for welding. Almost all link lengths for the loop method are too long, but the other methods have a spread more centered around the target value. The loop method is the easiest to accurately fabricate because sewing with a needle is the most representative of selecting a point on the vine body. Its large link length errors could be due to inaccuracies in modeling link length when there is a gap between the two points being connected.

The results of a statistical analysis of the measured DH parameters are shown in Figure \ref{fig:dh_stats}. We omitted data from the welded LDPE vine robots as they are unable to grow without material failure. We analyzed the independent contributions of fabrication method, material, and growth. We did not analyze the coupled effects of any of these factors because of limited sample size. We ran a one-way ANOVA on the effect of methods for link twist and link length data, resulting in p-values of 4e-06 and 2.5e-08, respectively. The joint data grouped by methods failed the homogeneity of variance assumption, p = 0.0225, so we ran a Kruskal-Wallis test on it instead, p = 0.0022. We then ran Tukey's honest significant difference test on each of the DH parameters to identify pairs of methods that are significantly different from each other. These significant pairs are indicated in Figure \ref{fig:dh_stats} with vertical lines and asterisks.

We ran an independent t-test on the materials for joint and link length data, resulting in p-values of 0.7028 and 0.1663 respectively. The twist data grouped by material failed the homogeneity of variance assumption, p = 0.0023, so we ran a Welch two-sample t-test on the materials data for this DH parameter, p = 0.0581.

Measurements pre-growth and post-growth are inherently linked. Therefore, we ran a dependent t-test to analyze the data for each DH parameter group by pre-growth and post-growth. Link twist and joint angles changed significantly before and after repetitive growths. The mean of link lengths change by 0.02 mm after repetitive growths, and this data has a p-value of 0.9474. This is an expected behavior because growth should not affect the location of the joints along the length of the vine. The non-zero difference in the mean is likely due to the fact that we cannot place the optical markers at exactly same location each time.